\documentclass[sigconf]{acmart}

\AtBeginDocument{%
  \providecommand\BibTeX{{%
    \normalfont B\kern-0.5em{\scshape i\kern-0.25em b}\kern-0.8em\TeX}}}

\usepackage[T1]{fontenc}
\usepackage{lmodern}
\begin{document}
\def\x{{\mathbf x}}
\def\L{{\cal L}}
\def\eg{\textit{e.g.}}
\def\ie{\textit{i.e.}}
\def\Eg{\textit{E.g.}}
\def\etal{\textit{et al.}}
\def\etc{\textit{etc}}
\newcommand{\blue}[1]{\textcolor{blue}{#1}}

\title{\emph{The eyes know it}: FakeET- An Eye-tracking Database to Understand Deepfake Perception}

\author{Parul Gupta}
\affiliation{%
  \institution{Indian Institute of Technology Ropar}}
	\email{2016csb1048@iitrpr.ac.in}

\author{Komal Chugh}
\affiliation{%
  \institution{Indian Institute of Technology Ropar}}
  \email{2016csb1124@iitrpr.ac.in}
	
\author{Abhinav Dhall}
\affiliation{%
  \institution{Monash University/Indian Institute of Technology Ropar}}
  \email{ahinav.dhall@monash.edu}
	
\author{Ramanathan Subramanian}
\affiliation{%
  \institution{Indian Institute of Technology Ropar}}
  \email{s.ramanathan@iitrpr.ac.in}

%
%
%
%
%
%

\renewcommand{\shortauthors}{Trovato and Tobin, et al.}

\begin{abstract}
We present \textbf{FakeET}-- an eye-tracking database to understand human visual perception of \emph{deepfake} videos. Given that the principal purpose of deepfakes is to deceive human observers, FakeET is designed to understand and evaluate the ease with which viewers can detect synthetic video artifacts. FakeET contains viewing patterns compiled from 40 users via the \emph{Tobii} desktop eye-tracker for 811 videos from the \textit{Google Deepfake} dataset, with a minimum of two viewings per video. Additionally, EEG responses acquired via the \emph{Emotiv} sensor are also available. The compiled data confirms (a) distinct eye movement characteristics for \emph{real} vs \emph{fake} videos; (b) utility of the eye-track saliency maps for spatial forgery localization and detection, and (c) Error Related Negativity (ERN) triggers in the EEG responses, and the ability of the \emph{raw} EEG signal to distinguish between \emph{real} and \emph{fake} videos.
\end{abstract} 




\maketitle

\section{Introduction}
The commonplace availability of image and video \emph{forgery} software has led to the widespread creation of image and video-based \emph{deepfakes} for purposes such as trolls, disinformation campaigns and political propaganda. Even if deepfakes may not completely mislead people, they nevertheless contribute to uncertainty and distrust of media content, posing a significant challenge to democratic societies~\cite{vaccari2020}. Therefore, artificial intelligence (AI)-based \emph{fake detection} (FD) techniques are critical for governments to inform citizens and shape public policy.

\begin{figure}[t]
\centering
  \includegraphics[width=\linewidth]{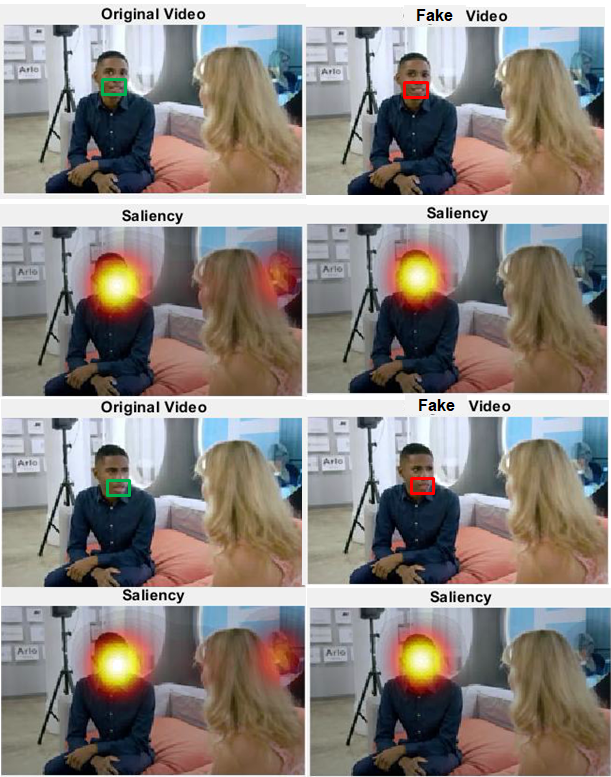}   
	\caption{\emph{Real} vs \emph{fake} video viewing patterns: Two frames and eye-track heatmaps are shown for a \emph{real} (left column) and a \emph{manipulated} video (right column). Original and forged mouth regions are shown via green and red rectangles. We observed an \emph{exploratory} viewing behavior for \emph{real} videos, and an \emph{explanatory} viewing pattern for \emph{fake} videos. This results in diffused gaze for \emph{real} vs compact gaze for \emph{fake} videos; \eg, visual attention to the onlooker and other scene characters can be noted for the \emph{real} frames, but is limited to the manipulated face in the \emph{fake} video frames. View in color and under zoom.} \vspace{-.2cm}
  \label{fig:moneyshot}
\end{figure}

While human performance has for long been the `gold standard' for AI methods to match and surpass, deepfakes are expressly designed with the objective of spoofing human detection. Indeed, computers can examine visual information that are inconspicuous to humans~\cite{pasquini2017statistical}. Therefore, it would be reasonable to envisage optimized FD via human-machine cooperation; \eg, information forensic experts could hypothesize probable forgeries in an image/video with relative ease, while state-of-the-art AI methods could follow-up to validate these hypotheses.  
While user impressions have been utilized to evaluate the efficacy of forgery techniques~\cite{FaceForensics++,jiang2020deeperforensics10}, user behavioral cues have never been utilized for FD to our knowledge. Specifically, we argue that \emph{implicit} user cues such as gaze patterns and neural responses can be very useful in this regard, as they can be acquired with minimal effort via wearable sensors. As evidence of this argument, we present \textbf{FakeET}\footnote{which would be made publicly available on paper acceptance}, a dataset of gaze and EEG recordings acquired from 40 users exposed to \emph{Google Deepfake}~\cite{DDD_GoogleJigSaw2019} videos.

Figure~\ref{fig:moneyshot} presents two exemplar frames from a \emph{real} and \emph{fake} video, and corresponding gaze heatmaps. The deepfake video includes forgeries localized to the mouth region, and the original and forged image regions are denoted via green and red rectangles. When users are posed with the \emph{fake video detection} task, they \emph{explore} the scene for visual irregularities while viewing a \emph{real} video; conversely, an \emph{explanatory} viewing behavior is observed when users infer a forgery candidate. Their subsequent visual processing is focused on verifying if the probable forged region is indeed different from its neighborhood. This results in diffused gaze patterns for \emph{real} videos, and more focused patterns for \emph{fake} videos.

The exploratory vs explanatory viewing patterns can be quantified via a number of eye track-based measures such as \emph{number of fixations} (more fixations on \emph{real} videos), \emph{scan-path length} (longer scan-paths on \emph{real} videos), \emph{repeated fixations} conveying propensity for scene exploring (fewer repeated fixations on \emph{fakes}) and \emph{fixation entropy} indicating consistency/stochasticity in visual processing (lesser entropy for \emph{fakes}). Evidently, these are useful user-centric features for the purpose of detecting deepfakes. Furthermore, neural triggers such as Error-Related Negativity (ERN), which occur when users become aware of erroneous information, are observable from EEG patterns acquired for \emph{fake} videos. Via experiments, we demonstrate that (a) gaze heatmaps can act as a reliable \emph{prior} by deep neural networks for examining forgeries, and (b) the EEG data acquired during \emph{real}/\emph{fake} video viewings can achieve better-than-chance FD performance. 

Overall, this work makes the following research contributions: (1) To our knowledge, FakeET represents the first compilation of user behavioral responses for deepfake detection; (2) Distinctive patterns characteristic of information irregularities can be observed for both the eye-gaze and EEG modalities, and (3) Through extensive experimental analysis, we show the utility of both the eye-gaze and EEG modalities in enabling automated deepfake detection.

\section{Related Work}
We review works that have (a) attempted FD by mining multimedia content, and (b) examined user impressions of synthesized deepfakes in this section.

\subsection{Content-based FD}
Since deepfakes have gained popularity in recent years, there has been increasing number of efforts in designing robust forgery detectors. Some of the initial works focused on inconsistencies exhibited in the physical or physiological aspects in the deepfake content. For example, \cite{visual_artifacts} used relatively simple visual aspects such as eye colour, missing reflections, and missing details in the eye and teeth areas. Similarly,  \cite{in_ictu_oculi} exploited the observation that many deepfake videos lack reasonable eye blinking due to the use of online portraits as training data. Detection systems based on head movements have also been proposed in the literature. Yang et al. in \cite{inconsistent_headposes} proposed that 3D head poses estimated from a face image can be used to detect errors introduced due to splicing synthesised face regions into the original image in deepfake creation. 

Some of the recent works have employed Deep Neural Networks (DNNs) for distinguishing between real and fake. In \cite{multi-task}, Nguyen et al. proposed a CNN system that uses multi-task learning to simultaneously detect fake videos and locate the manipulated regions. The network was based on an auto-encoder. Another interesting research by Li and Lyu \cite{Li2018ExposingDVFWA} hypothesised that manipulation techniques can create images of limited resolution, which need to be further warped to match the original faces in the source video which leaves artifacts. Hence, they use CNNs to detect artifacts from the detected face regions and the surrounding areas. Afchar et al. \cite{afchar2018mesonet} proposed MesoInc-4, an inception inspired \cite{Inception-v4} convolutional neural network with a small number of layers. A capsule network architecture is proposed in~\cite{Capsule_Forensics} which requires fewer parameters to train than very deep networks. 

\subsection{User-centered Evaluation}
The user studies conducted for the field of Deepfakes have aimed to evaluate human performance in the task of forgery detection. For instance, FaceForensics++ \cite{FaceForensics++} conducted the study as per-frame binary classification problem where each user was instructed to classify an image as real or fake in a limited time period. Different types of image qualities were used and the average human performance was observed to decrease with deteriorating quality.  DeeperForensics-1.0 \cite{jiang2020deeperforensics10}, another large-scale dataset asked the users about the realness of a video clip and provide rating on a scale of five. Video clips from various other datasets were included as well and the main purpose of this study was to examine the quality of their dataset as compared to others. Another interesting study which explains the contribution of deepfakes in spreading misinformation is that of \cite{ImpactOfDeepfakes}. The authors conclude that deepfakes sow uncertainty in the viewers' mind which in turn can lead to reduction of trust in news on social media platforms.

\subsection{Analysis of Related Work}
Based on the prior work, we note that hardly any user-centric FD methods exist; users have only been utilized to evaluate the efficacy of deepfakes. Most FD methods are purely AI driven, and focus on FD exclusively as a machine learning task. On the contrary, in this work, we explore the utilizing user-centric information for FD.  Also, we demonstrate FD using eye gaze and EEG information. To this end, FakeET is the first database containing \textit{implicit} human annotations as additional information (or additional user labels, similar to~\cite{Gilani16,wen2016}) for empowering machine-based FD methods. 

\section{User-centered Analytics}

This section describes (a) our user study, (b) differences in gaze patterns for \textit{real} vs \textit{fake} videos, (c) utility of the compiled eye-track saliency maps for FD, and (d) analysis of the EEG recordings. 

\begin{figure}[t]
    \centering
    \includegraphics[width=\linewidth]{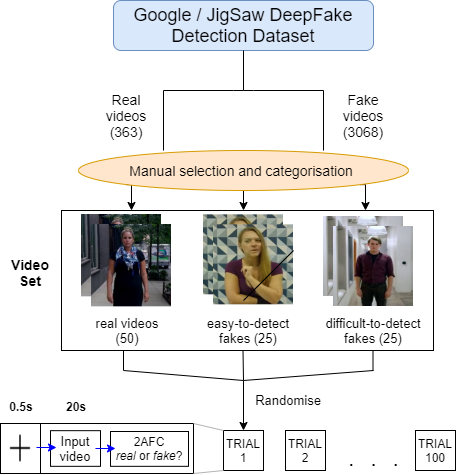}
    \caption{Overview of our user-study protocol.}
    \vspace{-.2cm}
    \label{fig:exp_protocol}
\end{figure}

\subsection{Materials \& Methods}

\subsubsection{Videos}\label{Vid_sel}
We used the Google/Jigsaw DeepFake Detection dataset \cite{DDD_GoogleJigSaw2019} for our user study. This dataset comprises 3068 \textit{fake} and 363 \textit{real} videos, out of which 811 videos (331 \textit{real}, 480 \textit{fake}) were used in our study. Furthermore, 
the \textit{fake} videos were categorized into \textit{easy} and \textit{difficult}-to-detect forgeries based on visual inspection; this taxonomy resulted in 244 \textit{easy} and 236 \textit{difficult}-to detect \textit{fake} videos.

\subsubsection{Participants}: 40 subjects (28 male, age 24.5 ± 4.3 and 12 female, age 22.8 ± 6.8) took part in our study. Most subjects were university graduate and undergraduate students, who were naive to the purpose of our study. All subjects had normal or corrected vision, and provided informed consent for participation as per the ethics approval guidelines. Participants were paid a nominal fee as a token of appreciation.  

\subsubsection{Stimuli}: We created a total of 10 video sets, with each set containing an equal number of \textit{real} and \textit{fake} videos, \ie, each set comprised 50 \textit{real}, 25 \textit{easy} and 25 \textit{hard}-to-detect fake videos. Given the fewer number of \textit{real} videos in the Google deepfake dataset, some \textit{real} videos were used across multiple sets. All videos had a resolution of $1920\times1080$ pixels, and a frame rate of 24 fps. All videos were clipped to 20 seconds duration for experimental purposes.

\subsubsection{Protocol}: Our experimental protocol was developed using the Matlab Psychtoolbox~\cite{psychtoolbox}. Each user viewed a set of 100 videos, and each set of videos was presented to 4 users in random order (10 video sets $\times$ 4 users/set = 40 users). 
Our experimental protocol is depicted in Fig.~\ref{fig:exp_protocol}. During each video presentation, termed as a \textit{trial}, a fixation cross was presented for 500ms, followed by the video playing for 20s. Upon video viewing, 
users were presented with a two-alternative forced choice (2AFC) task for recording their first impressions. Users had to label the video as \textit{real} or \textit{fake} via a radio button. As users viewed the videos, their eye-movements were captured via the \textit{Tobii TX300} eye-tracker with a sampling rate of 300 Hz. Also, their EEG responses were recorded with the 14-channel \textit{Emotiv Epoc}+ device, which captures EEG samples at 128 Hz frequency. The video presentation was synced with the eye-tracker and EEG recordings through the use of event markers. To minimize calibration
errors and user fatigue, the experiment was split into 4 parts comprising 25 trials each, interspersed with  relaxation breaks. All users took around 60 minutes to complete the experiment.

\begin{table*}[!htbp]
\small
\caption{FakeET in a nutshell.}\label{tab: Fake_ET}\label{ET_NS}\vspace{-.3cm}
\begin{center}
\begin{tabular}{|ccccc|}
\hline
\textbf{\# Videos} & \textbf{\# Users- ET} & \textbf{\# Viewings/video} & \textbf{\# Users- EEG} & \textbf{EEG description}\\ \hline
\textbf{811} (331 \emph{real}, 480 \emph{fake}) & \textbf{40} & \textbf{min}: 2, \textbf{max}: 16 & \textbf{30} & \textbf{\emph{Corrupt}} EEG epochs: 1236 \emph{real}, 1220 \emph{fake}\\
& & \textbf{mean:} $5\pm2.2$ & & \textbf{\emph{Clean}} EEG epochs: 636 \emph{real}, 793 \emph{fake}\\
\hline
\end{tabular}
\end{center}
\vspace{-.2cm}
\end{table*}

\begin{figure*}[!htbp]
\centering
    \includegraphics[width=0.24\linewidth]{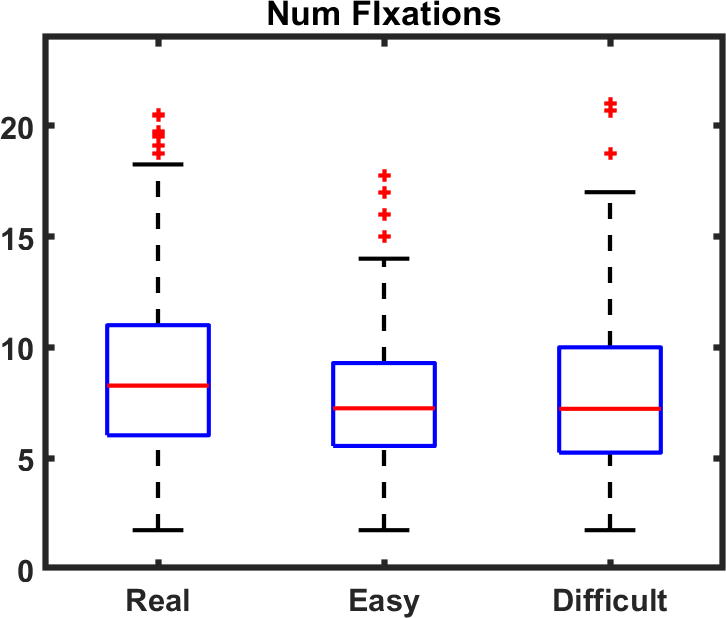}\hspace{0.05cm}\includegraphics[width=0.24\linewidth]{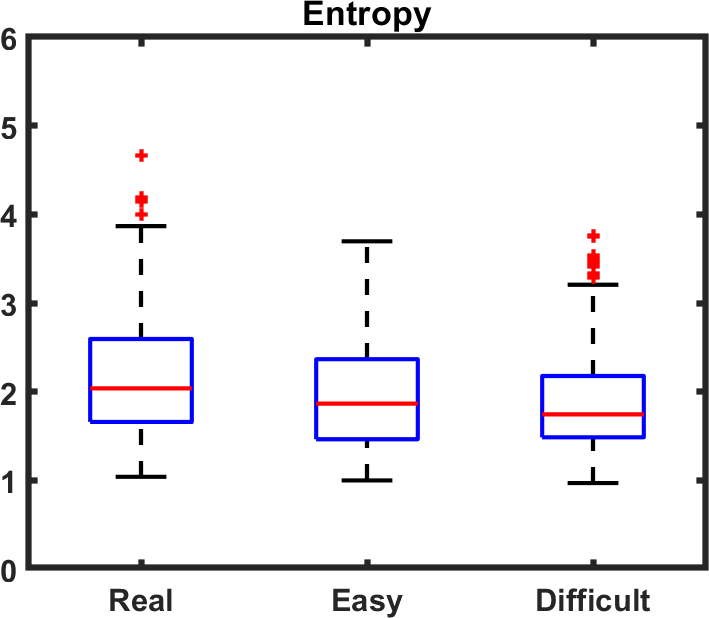}\hspace{0.05cm}\includegraphics[width=0.25\linewidth]{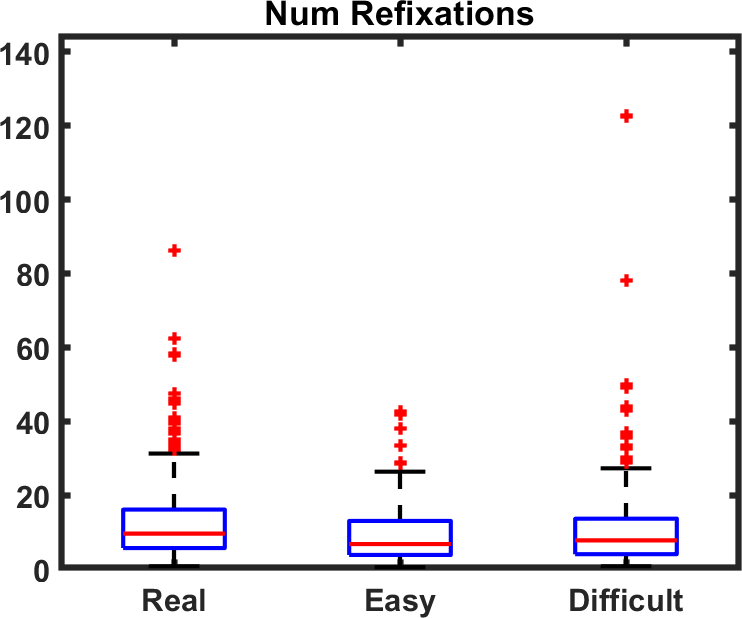}\hspace{0.05cm}\includegraphics[width=0.24\linewidth]{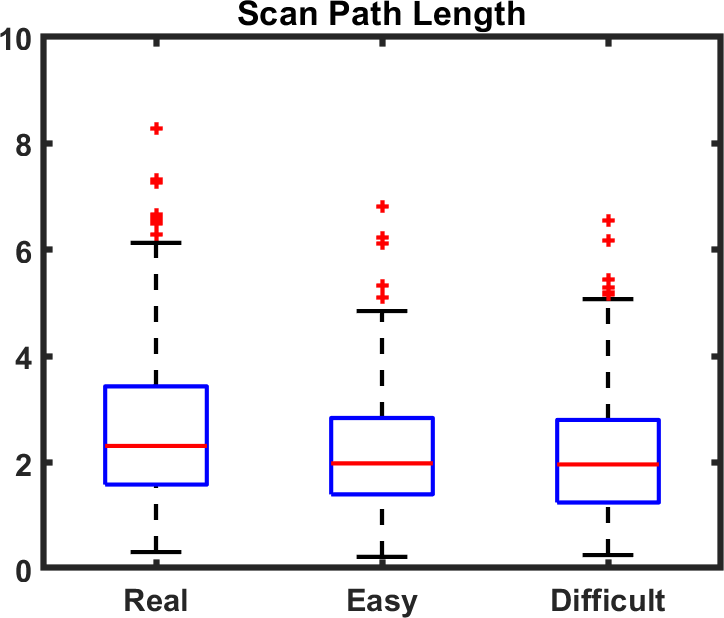}\vspace{-.2cm}
    \caption{Comparing viewing patterns differences among \emph{real}, \emph{easy} and \emph{difficult}-to-detect \emph{fake} videos via eye-track measures.}
    \vspace{-.2cm}
    \label{fig:et_comp}
\end{figure*}

\subsection{Gaze Pattern Analysis}\label{ga}

Table~\ref{ET_NS} summarizes statistics of the FakeET dataset, in terms of the eye-gaze and EEG recordings available for research. While a substantial amount of EEG data was visually found to be noisy, and ignored for our analyses, we nevertheless note that sophisticated machine learning algorithms designed for noisy data (\eg, multiple instance learning, Siamese neural networks) can be explored to improve EEG-based FD as in~\cite{Shukla20}. In terms of eye-gaze recordings, each video corresponds to a minimum of 2, and a maximum of 16 eye-track files (some data was lost due to subjects closing their eyes, tracker errors, \etc.). The following sections demonstrate the validity of the compiled eye-gaze and EEG data.

We first examined if the compiled gaze recordings (a) were distinctive for \textit{real} and \textit{fake} videos, so that they could be utilized as features for FD, and (b) if employing the eye-track heatmap as a \textit{prior} for deep neural networks optimized FD performance.

\subsubsection{Statistical Analyses}\label{et_sa}

As seen from Figure~\ref{fig:moneyshot}, the \emph{explorative} vs \emph{explanative} user viewing patterns for \emph{real} vs \emph{fake} videos can be captured via a number of low-level eye-gaze measures. To this end, upon processing the raw gaze data, and discovering fixations and saccades via the EyeMMV toolbox~\cite{Krassanakis_Filippakopoulou_Nakos_2014}, we computed the following measures: 
\begin{itemize}
\item[(1)] \textbf{Mean number of fixations:} falling on 20s of video. More fixations should indicate greater scene exploration.
\item[(2)] \textbf{Mean fixation duration:} over each video. Longer fixation durations indicate visual processing by the users to assimilate and understand the scene content. 
\item[(3)] \textbf{Fixation entropy:} conveying the randomness with which viewers fixate on scene regions. Entropy values can theoretically vary from [$0,\infty$], with 0 denoting that every viewer fixates on the same exact scene location, while $\infty$ denotes absolute viewing stochasticity. Intuitively, presence of visual attractors in the scene, such as a manipulated scene region, should result in lower viewing entropy, while scene explorations are correlated with high entropy. 
\item[(4)] \textbf{Number of re-fixations:} Computing the number of recurring fixations within a scene region, and 
\item[(5)] \textbf{Scan-path length:} Length of user scan paths over the video scene. Longer scan paths indicate explorative viewing. 
\end{itemize} 

We then compared these measures for the \emph{real}, \emph{easy} and \emph{difficult}-to-detect fake videos. As noted in Sec.~\ref{Vid_sel}, \emph{fake} videos were further categorized into \emph{easy} and \emph{difficult}-to detect fakes by two of the paper authors. This was owing to the inconspicuous manipulations existing in some videos, and since our user-study was performed with naive viewers, we hypothesized different user responses for the \emph{easy} and \emph{difficult} fakes. We posited that if users had greater difficulty in detecting the difficult manipulations, there should be significant differences in viewing patterns for the \emph{easy} and \emph{difficult} fakes.
 
Fig.~\ref{fig:et_comp} compares viewing patterns for the \emph{real} vs \emph{easy} and \emph{difficult} fakes. A one-way Analysis of Variance (ANOVA) test was performed to identify significant differences in eye-track measures. We noted the \emph{main effect} of (real/easy/difficult) viewing condition (with $p<0.05$) for the \emph{mean number of fixations} (more fixations on \emph{real} videos), \emph{entropy} (greater viewing entropy for \emph{real} videos), \emph{number of re-fixations} (more re-fixations on \emph{real} videos) and the \emph{scan-path length} (longer scan-paths on \emph{real} videos) measures, confirming that \emph{\textbf{viewing differences existed for real vs fake videos}}. We then examined differences for \emph{easy} vs \emph{difficult}-to-detect fakes via the two-sample $t$-test. This analysis revealed no differences, except for the \emph{number of re-fixations} measure, implying that \emph{\textbf{detecting difficult fakes entailed a greater amount of visual processing}}. There were however, no major differences in the mean fixation durations over the three conditions. Overall, extracting eye-gaze statistics revealed (a) the existence of differences in eye-gaze metrics for \emph{real} vs \emph{fake} videos, implying that eye-gaze features \emph{are discriminative} for FD, and (b) not much differences existed for \emph{easy} vs \emph{difficult} fakes, except for a greater degree of visual processing on \emph{difficult fakes} for inference.

\subsubsection{FD via eye-track heat maps}\label{et_exp}
As observed in our study, the user gaze pattern is more spatially spread across the scene in case of a real (un-manipulated) video. On contrary in the case of a fake (manipulated) video, the user's gaze pattern is spatially focused around the face area only. Grimes et al.\ \cite{Grimes1991MildAD} conducted an interesting study to investigate, how the subjects' attention is divided between the audio and visual channels, while watching television news. In one of their experiments, they introduced manipulations in the visual channel, and found that the subject's attention increased. This could be attributed to manipulated region being more salient to the human visual system. In other words, the subject paid more \emph{attention} when the videos had an unusual visual channel, as compared to the \emph{real} un-manipulated videos. This is similar to the observation regarding Fig.~\ref{fig:moneyshot} in our study. The fake videos have some perturbations, which can be mostly observed in the visual channel, thus diverting the users' gaze (or attention) towards such regions. To empirically validate these observations, we trained a CNN to predict if a video is \emph{real} or \emph{fake}. 

The base of the CNN in our study is a standard 3D ResNet \cite{DBLP:journals/corr/abs-1708-07632}. Table \ref{tab:stream_arch} summarises the network structure. We trained the network in three different settings based on the following network inputs: (a) full-(video) frame ; (b) full-frame overlaid by user gaze maps, and (c) the cropped face region only. In the first setting, network is input with the full frame. In the second case, we overlay the full frame with the binarized gaze pattern map. The rationale behind adding the gaze pattern map is to augment information pertaining to \emph{important} (or equivalently, possible forged) scene regions from a user perspective. In the third scenario, we crop the facial region using S\textsuperscript{3}FD face detector ~\cite{zhang2017s3fd} to pass the detected face region as the input to the network. The third scenario is designed on the observation that all available deepfake datasets contain video manipulations limited to the facial region, and therefore, the machine can assume \emph{a-priori} that the face needs to be mined for visual/statistical irregularities.

\begin{figure}[t]
    \centering
    \includegraphics[width=\linewidth]{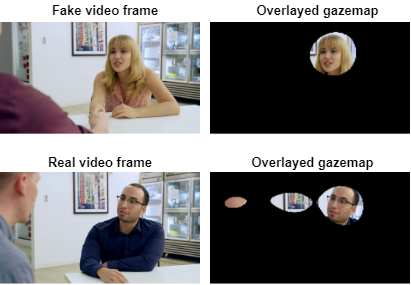}
    \caption{Different CNN inputs (Section \ref{et_exp}): Top and bottom rows show frames from fake and real video, respectively. Left column shows input to the CNN, when full-frame is used. Right column shows input to CNN, when the gaze map is overlaid on the original frame (left column). Note that these regions, where the image content is visible, represent the parts of image, where the subjects in the study fixated.}
    \vspace{-.2cm}
    \label{fig:object_focus_in_real}
\end{figure}

\begin{table}[b]
  \begin{tabular}{|c|}
    \hline
    \textbf{CNN Structure} \\ \hline
    conv1 \\ \hline
    conv2\_x\\ \hline 
    conv3\_x\\ \hline
    conv4\_x\\  \hline
    conv5\_x\\   \hline
    average pool\\  \hline
    fc7, 256$\times$7$\times$7, 4096\\  \hline
    batch\_norm\_7, 4096\\ \hline 
    fc8, 4096, 1024\\ \hline
    batch\_norm\_8, 1024\\  \hline
    dropout, $p=0.5$\\ \hline
    fc10, 1024, 2\\ \hline
  \end{tabular}
	\vspace{.1cm}
  \caption{Structure of the CNN architecture (initial layers are the same as in the 3D ResNet architecture \cite{DBLP:journals/corr/abs-1711-09577}).}
  \label{tab:stream_arch}
\end{table}

\textbf{Method: } The dataset $D = \{(v^1,y^1), (v^2,y^2), ... , (v^N,y^N)\}$ consists of $N = 805$ videos (331 real and 474 fake), and we split it into $60:20:20$ train-validation-test sets. Here, $v^i$ denotes the input video and the label $y^i \ \epsilon \  \{0,1\}$ indicates whether the video is \emph{real} ($y^i=1$) or \emph{fake} ($y^i=0$). To encode the temporal changes at a finer level, we divide each video into $D$-second long segments - $\{s^i_1, s^i_2, ... , s^i_n\}$, where $n$ denotes segment count for an input video $v^i$. Each segment either contains the chunk of full frames, or cropped faces only, or the full frames masked with the binarised gazemaps, depending upon the setting under which the network has to be trained. We use \emph{cross-entropy loss} to train our model, whose input is $s^i_t$, a video sequence of size ($C \times h \times w \times D \times f$), where C ($=3$) refers to the RGB color channels of each frame, $h, w$ are the frame height and width and $f$ is the video frame rate and D is the duration of each chunk (D=1 in our implementation).

Fig. \ref{fig:object_focus_in_real} shows sample frames from  \textit{real} and \textit{fake} videos along with user gaze maps overlaid on them. The first column is an illustration of the input when the network is trained with full-frame information (which includes a great deal of background \emph{noise}), while the second column is for the scenario when the network takes in frame information augmented with eye gaze/attention masks. As it can be observed from the figure, the gaze map for a \textit{fake} sample is centered around the manipulated face only whereas for a \textit{real} sample, the user's gaze also explores the objects around a person to verify the naturalness of a scene.\\

\textbf{Results: } The video-wise \textit{area under the curve} (AUC) metric was used for evaluation. Table \ref{tab:results} shows the comparison of the three settings. It is observed that full-frame input achieved 49.03\% ± 0.83. When the frame information is overlaid with gaze-based masks derived from the user study, the AUC increases to 54.39\% ± 0.01. This validates the observation that with the gaze maps, more focused information is input to the network. This can also be related to recent concept of attention layers in CNN. Instead of explicitly adding attention layer, we are adding gaze maps generated through user study. This sets the scene for a collaborative effort for FD between human and machines by bringing humans-in-the-loop.

The cropped faces gives a higher video-level AUC score = 61.75\% ± 1.25. The is better than both the full-frame version of cropped faces as well as the gazemaps used as masks. This can be attributed to the fact that while the gaze is likely to fall on the manipulated facial region, it may also fall on some background pixels, inherently due to eye movements or tracker errors. Further, it is important to note that in the current datasets, manipulations are localized to the face region. 

With evolutions in deepfake synthesis, manipulations need not necessarily be performed on faces, and may involve other scene regions as well (\eg, person's hands, clothing, scene background, \etc.). In such situations, the face \emph{prior} assumption would fail, but gaze information could still convey such irregularities, as human eyes are sensitive to `center-surround' differences in the video content. Therefore, the gaze cue influences the CNN to \emph{focus} on certain scene regions and thereby improving FD performance; even if the focus may not be achieved as perfectly as with face crops, the eye-gaze prior is more generalizable than the face prior as deepfakes evolve.   

\renewcommand{\arraystretch}{1.5}
\begin{table}[t]
    \centering
    \begin{tabular}{|c | p{9em}| c|}
    \cline{1-3}
    \textbf{Full-frame} & \centering \textbf{Full-frame overlaid with gaze map} & \textbf{Face-only}\\
    \cline{1-3}
    49.03 ± 0.83 & \centering 54.39 ± 0.01 & 61.75 ± 1.25 \\
    \cline{1-3}
    \end{tabular}
    \caption{Comparison of performance of CNN (Section \ref{et_exp}) based on different inputs. Note the increase in the performance, when the gaze maps are added to the full-frame input.} \vspace{-5mm}
    \label{tab:results}
\end{table}

\subsection{EEG Data Analysis}
As viewers attempted to detect \emph{fake} videos, their cognitive responses were recorded via an \emph{Emotiv} EEG headset. \emph{Emotiv} is a commercial and portable EEG device which comprises 14 electrodes, has a 128 Hz sampling rate, and is typically used for gaming applications; however, it has recently been used to successfully analyze user cognitive responses with a high degree of accuracy~\cite{Vi2014,Shukla20,Bilalpur17}. Due to erroneous and corrupted recordings, we could only use EEG data for 30 of the 40 users who participated in our study.

Corresponding to each \emph{trial}, \ie, video presentation timeline, we extracted \emph{epochs} of 10.5 seconds (0.5s pre-stimulus for baseline power subtraction plus 10s of stimulus viewing). This epoch time-length was deemed sufficient based on eye-track findings; each epoch was therefore represented by a 14 (EEG channels) $\times$ 1344 (10.5s $\times$ 128 Hz) matrix. Baseline subtraction, EEG band-limiting to [0.1, 45] Hz, visual epoch removal and Independent Component Analysis (ICA) were performed to eliminate corrupted epochs along with head, eye and muscle-movement artifacts. The cleaning  resulted in 1429 \emph{clean} epochs of which 636 arose from correctly user-labeled \emph{fake} video trials, and 739 accurately labeled \emph{real} video trials. Experiments were performed on this \emph{clean} EEG data. \\ 

\subsubsection{{ERP analysis:}} Event-related potentials (ERPs) denote the \emph{average} time and phase-locked neural response following an event of interest; \eg, P300 denotes an ERP typically elicited 300ms following presentation of faces. Error related Negativity (ERN) is an ERP triggered when the user becomes aware of obvious error(s) committed by self, peers or with the presented information. Authors of~\cite{Vi2014} successfully captured the ERN pattern, characterized by a negative peak followed by a positive peak, when the user becomes aware of an error committed by a collaborator during an interactive task. 

\begin{figure*}
\centering
  \includegraphics[width=0.32\linewidth]{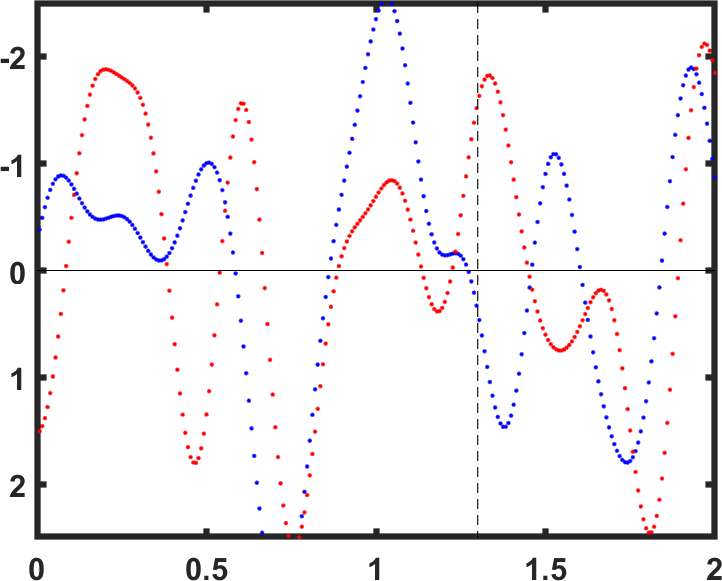}\hspace{0.02cm}\includegraphics[width=0.32\linewidth]{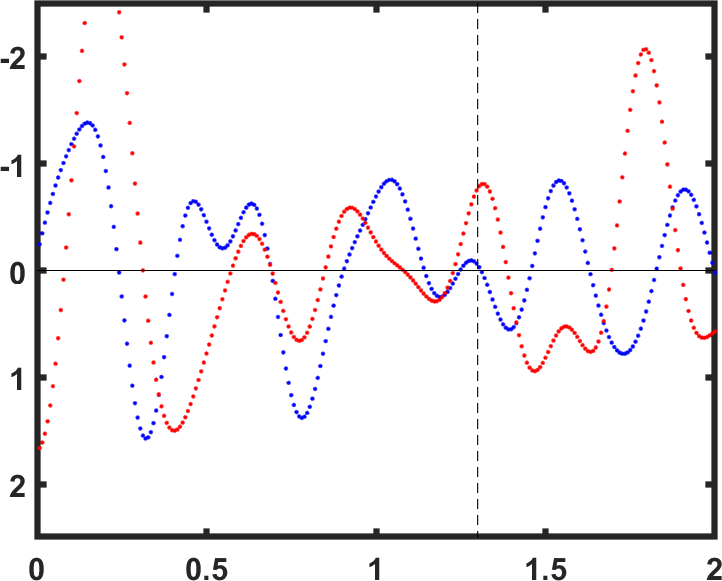}\textbf{\hspace{0.02cm}\includegraphics[width=0.32\linewidth]{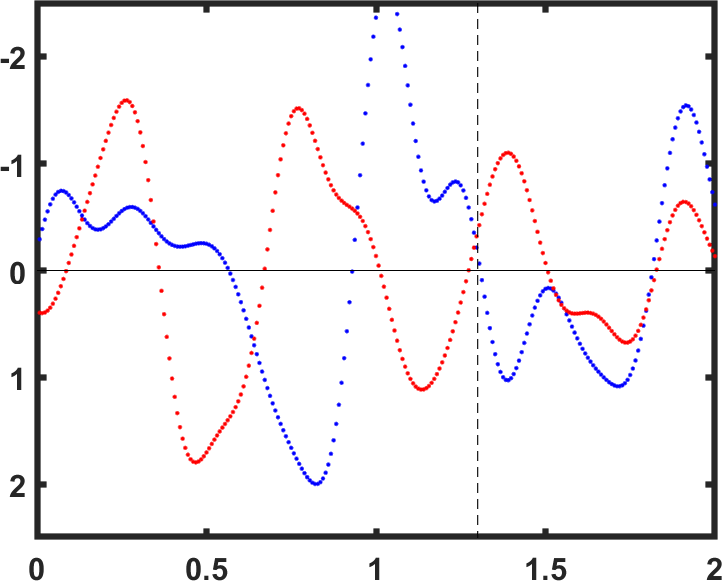}}
\caption{ERPs for the \emph{real} (blue) vs \emph{fake} (red) EEG response at the \emph{f7} (left), \emph{af3} (middle) and \emph{fc5} (right) electrode channels. An ERN pattern (-ve peak followed by +ve peak) can be observed around 1.3s post video onset. $x$ denotes time (s), while $y$ denotes potential ($\mu V$). ERPs are plotted upside down as per convention.}\vspace{-.1cm}
  \label{fig:EEG_ERPs}
\end{figure*}

Figure~\ref{fig:EEG_ERPs} presents ERPs associated with the \emph{f7}, \emph{af3} and \emph{fc5} \emph{Emotiv} electrodes, and indicate neural activity in the frontal-central part of the brain. ERPs ($y$-values) are plotted upside down as per convention, with the blue and red curves denoting ERPs for \emph{real} and \emph{fake} videos respectively. One can clearly observe an ERN-like pattern in the \emph{fake} ERPs around 1.3s (dashed vertical line) post video onset. Note that this pattern is inconspicuous for the \emph{real} ERPs. We attribute this $\approx$~1.3s latency to the video content; most videos in the Google Deepfake~\cite{DDD_GoogleJigSaw2019} involve a $\approx$~1s \emph{establishing shot}, typically comprising a camera pan or zoom or actor motion, to introduce the character(s) of interest . Once the scene is understood by the viewer, cognitive inference of anomalous scene regions appears fairly instantaneous. ERP findings (a) confirm distinctive patterns in user cognitive responses upon visual processing of fake videos, and (b) validate the utility of the compiled EEG data.  

\subsubsection{EEG-based FD}
We determined the efficacy of the EEG modality for FD via classification experiments on the \emph{clean} EEG data via following baselines. \\

\textbf{Majority label assignment (Maj-Class):} Where every test sample is assigned the label of the majority class in the training set.

\textbf{Random label assignment (Random):} A random prediction score \emph{s} $\in$ [0,1] is assigned to the test sample, and the test label \emph{l} is assigned as $l = s > 0.5$.

\textbf{Naive Bayes (NB):} Known as the \textit{minimum risk} classifier, NB assigns the test label based on the maximum \emph{a-posteriori} (MAP) criterion. Despite its simplicity, NB is known to achieve competitive performance for many real world problems. 

\textbf{Logistic:} Assigns a score \emph{s} $\in$ [0,1] as $s = \frac{1}{1+e^(-w^Tx)}$, where $x,w$ denote the test feature and a vector of learned regression weights respectively.
  
\textbf{$k$-Nearest neighbors ($k$-NN):} Where a test sample is assigned the class label representing the membership of the majority of its $k$ nearest neighbors. We found $k=3$ to achieve optimal classification performance.

\textbf{Decision tree (Dec-Tree):} Which represents a rule-based classifier to predict outcomes; class labels are determined at the terminal/leaf nodes of the tree. 
  
\textbf{Linear SVM (Lin SVM):} which represents a very powerful classifier for small-to-medium scale datasets. SVM performance was fine-tuned by varying the mis-classification tolerance parameter \emph{C} within [$10^{-6},10^6$]. 

\textbf{Time-series CNN (T-CNN):} In addition to the above shallow learners, we also applied a 3-layer deep convolutional neural network (CNN), comprsing two 1-D convolutional layers followed by a fully-connected layer, employed for EEG data analysis in~\cite{Shukla20}.

\textbf{Results and Discussion:} As over-fitting is a common problem where few high-dimensional training examples are available (we had 1429 epochs, each of which was of $14\times10344$ size), we first applied Principal Component Analysis to retain 95\% data variance upon vectorizing the epoch data; this reduced epoch dimensionality to 679. PCA-ed data was input to each of the classification algorithms, and the EEG-based FD results are presented in Table~\ref{tab:EEG_scores}. Along with the AUC measure which is employed for evaluation in Section~\ref{et_exp}, we also computed the F1-score (harmonic mean of precision and recall), and the accuracy measure for benchmarking. We also performed 20 repetitions of five-fold cross-validation (total of 100 runs), to examine generalizability of the results.  

Since AUC computation requires classifier prediction scores (probabilities), AUC scores are not available for the Maj-Class, \emph{k}-NN and Dec-Tree methods. Due to the imbalanced nature of our EEG data where the \emph{fake} epochs are in minority, majority class assignment results in an accuracy of 0.56, equal to the \emph{real} class proportion. However, Maj-Class achieves zero recall for the \emph{fake} class, resulting in an F1-score of 0. The Random classifier expectedly achieves an accuracy of 0.5, AUC of 0.5, and a slightly lower F1-score of 0.47. Cumulatively, the simplistic Maj-Class and Random classifiers convey that (a) the F1-score is more sensitive to class imbalance than AUC, and (b) classifiers that work better on the minority \emph{fake} class are more likely to achieve a higher AUC/F1-score. 

To benchmark the performance of other classifiers, we compared their AUC and F1-score distributions  against the Random outputs via the two-sample $t$-test. Starred values in Table~\ref{tab:EEG_scores} denote distributions significantly higher than the Random classifier. From the observed results one can conclude that (a) Logistic regression and linear SVM perform similar to a random classifier, (b) the Dec-Tree and T-CNN algorithms achieve a significantly higher F1-score than Random, (c) the $k$-NN classifier achieves the second-highest F1 metric of 0.54, and (d) the Naive Bayes classifier achieves optimal EEG-based FD with AUC and F1 scores of 0.55. 

Cumulatively, EEG-based classification results reveal that (1) Better-than-chance FD is achievable only using PCA-ed EEG features and off-the-shelf classification algorithms, and (2) While a modest FD performance is achieved in this work, better EEG descriptors (Power Spectral Density features, EEG spectrograms, \etc) and superior classification strategies may be explored for possible improvements.

\begin{table*}[!tbph]
  \caption{Classification performance with different algorithms on EEG epochs (+ve class proportion = 636/1429 = 0.445).All scores are computed over 20 repetitions of 5-fold cross-validation (total 100 runs). AUC/F1 distributions significantly higher than the random output (with $p<0.05$), as determined via a 2-sample \textit{t}-test, are denoted by *.}\vspace{-.1cm}
  \label{tab:EEG_scores}
  \scalebox{0.98}{
  \begin{tabular}{|c|c|c|c|c|c|c|c|c|}
    \cline{1-9}
    \textbf{Classifier} & \textbf{Maj-Class} & \textbf{Random} & \textbf{NB} & \textbf{Logistic} & \textbf{\emph{k}-NN} & \textbf{Dec-Tree} & \textbf{Lin SVM} & \textbf{T-CNN}  \\
    \cline{1-9}
	   \textbf{Acc} & 0.56$\pm$0.00 & 0.50$\pm$0.03 & 0.52$\pm$0.02 & 0.50$\pm$0.02 & 0.52$\pm$0.03 & 0.52$\pm$0.03 & 0.51$\pm$0.03 & 0.51$\pm$0.03\\
		  \textbf{AUC} & - & 0.50$\pm$0.04 & \ \textbf{0.55$\pm$0.03*} & 0.50$\pm$0.02 & - & - & 0.51$\pm$0.03& 0.51$\pm$0.03\\
			\textbf{F1}  & 0.00$\pm$0.00 & 0.47$\pm$0.03 & \ \textbf{0.55$\pm$0.02*} & 0.45$\pm$0.03 & \ 0.54$\pm$0.04* & \ 0.51$\pm$0.07*& 0.46$\pm$0.04& \ 0.50$\pm$0.03*\\
	\cline{1-9}
\end{tabular}}
\end{table*}

\section{Discussion}
FD techniques have been primarily focusing on face manipulation in videos. As manipulation techniques progress in the future, it is going to be non-trivial for users to identify fakes. In our experiments, we observe that the EEG and gaze maps show that the \emph{explanatory} behavior is observed, when a user becomes aware of any irregularity in the video (similar to prior studies~\cite{Subramanian14,Shukla18}, where emotional content was found to be a strong attractor of visual attention). This opens up a new direction for the effort of detecting manipulated video using \emph{human-machine cooperation}. The research question would then be: \emph{Based on implicit user cues acquired while watching a video, can an AI-based FD technique be used to validate (with explanations) if the video is fake?} 

The AI based tool can further make a more informed decision using the video content and the implicit user feedback, as compared to only mining the video signal. This is important as in the future, manipulations will be added to not just the face but to the scene background, audio information and other attributes such as clothes of video subject(s), scene lighting, \etc. These complex manipulations would be a direct consequence of evolving video manipulation techniques. Therefore, AI methods alone may be inadequate due to training data bias, and the difficulty in creating varied and representative training sets, which can simulate the whole gamut of video manipulations. In this regard, a \emph{human-in-the-loop} appears a promising direction for FD, and the exploitation of contextual information in the form of human eye-gaze and EEG annotations, which can be acquired in a facile manner, would enable the AI system to optimize decision-making.

\section{Conclusion \& Future Work}
In this work, we present the first study that explores the use of user behavioral cues for fake detection. A novel dataset-- \emph{FakeEt} containing 811 samples from 40 subjects is curated via a user study, and contains two modalities: \emph{eye gaze} and \emph{EEG} signals. The gaze maps show an interesting spread of fixations outside of the face area(s) for \emph{real} videos. This exploratory gaze behavior is opposite to the explanatory one, in the case of \emph{fake} videos. Subjects explored the scene less in fake videos and fixated more around the face, which is the typica; region of manipulation in the video. Augmenting input frames with gaze maps, while training a CNN for fake detection, shows that the network implicitly learns from the gaze patterns and the performance improves by approximately 5\%. Similar observation is also made in the case of EEG signal analysis that a subject's EEG responses are discriminative for predicting if the viewed video is fake.

As future work, we will explore: (a) strategies for human-machine collaboration for deepfake detection; (b) learning a joint embedding to capture similarities between EEG and eye gaze patterns; (c) transfer learning from saliency prediction networks to gaze map generation, for increasing the number of data samples and better scene understanding; (d) explore fusion techniques for adding eye gaze maps through embeddings into the CNN input, and (e) analyze the effect of audio manipulation on EEG signals.

\bibliographystyle{ACM-Reference-Format}
\bibliography{sample-base}

%
%
%
%
%
%
%
%

\end{document}